\documentclass{bmvc2k}

\title{Enhancing Cardiovascular Disease Prediction through Multi-Modal Self-Supervised Learning}

\addauthor{Francesco Girlanda}{fgirlanda@student.ethz.ch}{1}
\addauthor{Olga Demler}{odemler@bwh.harvard.edu}{1,2}
\addauthor{Bjoern Menze}{bjoern.menze@uzh.ch}{3}
\addauthor{Neda Davoudi}{neda.davoudi@ai.ethz.ch}{1,3,4}

\addinstitution{
 Department of Computer Science\\
 ETH Zürich\\
 Zürich, Switzerland
}
\addinstitution{
 Brigham and Women’s Hospital\\
 Harvard Medical School\\
 Boston, Massachusetts, USA
}
\addinstitution{
 Department of Quantitative Biomedicine\\
 University of Zürich \\
 Zürich, Switzerland
}
\addinstitution{
 ETH AI Center\\
 ETH Zürich\\
 Zürich, Switzerland
}

\runninghead{Girlanda et al.}{Enhancing CVD Prediction through Multi-Modal SSL}

\def\etal{\emph{et al}\bmvaOneDot}

\usepackage[T1]{fontenc}
\usepackage{pdfpages}
\usepackage{url}
\usepackage{graphicx}	
\usepackage{hyperref}
\usepackage{amsmath} 
\usepackage{amsfonts}
\usepackage{svg}
\usepackage{multicol}
\usepackage{bbm}
\usepackage{authblk}
\usepackage{blindtext}

\usepackage{graphicx}
\usepackage{caption}
\usepackage{amssymb}
\usepackage{pifont}
\newcommand{\cmark}{\ding{51}}%
\newcommand{\xmark}{\ding{55}}%
\begin{document}

\maketitle

\begin{abstract}
    
Accurate prediction of cardiovascular diseases remains imperative for early diagnosis and intervention, necessitating robust and precise predictive models. Recently, there has been a growing interest in multi-modal learning for uncovering novel insights not available through uni-modal datasets alone. By combining cardiac magnetic resonance images, electrocardiogram signals, and available medical information, our approach enables the capture of holistic status about individuals' cardiovascular health by leveraging shared information across modalities. Integrating information from multiple modalities and benefiting from self-supervised learning techniques, our model provides a comprehensive framework for enhancing cardiovascular disease prediction with limited annotated datasets.

We employ a masked autoencoder to pre-train the electrocardiogram ECG encoder, enabling it to extract relevant features from raw electrocardiogram data, and an image encoder to extract relevant features from cardiac magnetic resonance images. Subsequently, we utilize a multi-modal contrastive learning objective to transfer knowledge from expensive and complex modality, cardiac magnetic resonance image, to cheap and simple modalities such as electrocardiograms and medical information. Finally, we fine-tuned the pre-trained encoders on specific predictive tasks, such as myocardial infarction. Our proposed method enhanced the image information by leveraging different available modalities and outperformed the supervised approach by 7.6\% in balanced accuracy. 
\end{abstract}

\section{Introduction}

Cardiovascular diseases (CVDs) are the leading cause of death worldwide resulting in 20.5 million deaths in 2021 \cite{lindstrom2022global} which makes the early diagnosis and prevention of CVD a critical task in medical research. 
Patient data, including laboratory values, imaging data, medical histories from the electronic health record (EHR), and lifestyle contribute to a comprehensive patient profile and are essential for healthcare decisions. Integrating these diverse data sources in real time facilitates more effective prevention and treatment strategies. Several medical data are used for detecting CVD. Cardiac magnetic resonance (CMR) images are the gold standard for assessing cardiac structure, and electrocardiograms (ECGs) are commonly used for electrophysiological evaluation. Early algorithms focus on automated ECG interpretation \cite{noseworthy2022artificial, khurshid2022ecg} or CMR image analysis \cite{raisi2022estimation}. A major limitation of current models is that they are limited to algorithms trained on a single modality and don't translate between different modalities, however, to capture the complexity of human biology, there is a need to go beyond traditional clinically-focused and expert-curated features and include critically important, but often neglected, data types on which doctors' evaluation rely on \cite{gong2022supervised}. Machine learning models can leverage the complementary information present in different modalities to develop a joint characterization of physiological states similar to the conventional approach of diagnosis by doctors and further enhancing their effectiveness. Several studies have employed multimodal data to improve diagnostic capabilities. Khurshid \etal developed a deep learning-based model to estimate left ventricular (LV) mass from 12-lead ECGs, which showed a strong correlation with LV mass on cardiac MRI imaging and cardiovascular events \cite{khurshid2021deep}. Christensen et al. developed a model that learned to interpret cardiac ultrasound images by correlating them with expert cardiologists' annotations \cite{christensen2024vision}. 
Borsos et al. proposed a multimodal deep learning architecture combining tabular data and imaging to predict dichotomized mRS scores three months post-Acute Ischemic Stroke \cite{borsos2024predicting}. Amal et al. provided a comprehensive review of multimodal approaches in healthcare for CVD analysis, emphasizing various fusion strategies \cite{amal2022use}. Ensemble learning improves deep learning model performance through various approaches but it is challenging due to high training costs, the need for inducing diversity among models, complex model selection, and combining predictions, requiring further research to optimize these aspects. \cite{ganaie2022ensemble}

Recent models have proven to be fruitful for use in biomedical prediction tasks, but the risk of overfitting remains due to the limited size of annotated datasets for supervised learning. Results of generative models such as Autoencoders (AEs) \cite{alain2014regularized} on multi-modal clinical measurements show that they perform well on aligning the embeddings from diverse modalities and constructing a holistic representation for characterizing physiological state \cite{radhakrishnan2023cross}. AEs are employed to learn cross-modal representations from large multimodal datasets. Healthcare datasets such as UK Biobank \cite{sudlow2015uk} serve as an excellent resource for learning clinically relevant representations. Hager \etal attempted to combine images and tabular data for multi-modal pre-training of representation by optimizing a CLIP loss and predicting myocardial infarction (MI) for the downstream task \cite{BestofBothWorlds}.They have shown that multi-modal prediction using CMR and tabular data outperforms uni-modal prediction. However, it relies solely on the image encoder for downstream tasks, ignoring the extensive information contained in the tabular data or other relevant modalities such as ECG.

In healthcare, ECG signals and basic medical information, are readily available or easily measured. However, CMR imaging, which provide detailed anatomical and physiological information about the heart, is costly to obtain and may not always be accessible. Turgut \etal tried a self-supervised contrastive learning \cite{chen2020simple} approach that transfers domain-specific information from CMR images to ECG embeddings. They predict the subject-specific risk of various CVDs and determine distinct cardiac phenotypes solely from ECG data and demonstrate that learned ECG embeddings incorporate information from CMR image with self-supervised training on both modalities \cite{turgut2023unlocking}. However, integrating tabular data, which includes demographics, lifestyle, and lab tests, with imaging data in multi-modal datasets is essential for informed clinical decision-making in healthcare \cite{acosta2022multimodal}. This project aims to improve the existing classification of CVD using only ECG signals and patient medical information for diagnosing MI by incorporating the embedding that is constructed with multiple modalities including CMR images.
This study demonstrates – to the best of our knowledge, for the first time – how to successfully integrate all available modalities in different formats as in realistic medical scenarios for predicting different downstream tasks. Furthermore, it improves the alignment of different modalities in latent space by leveraging the huge amount of unannotated data with self-supervised learning. Importantly, we increase the balanced accuracy by 7.6\% on a well-characterized clinical dataset for MI diagnosis.
Our code is publicly available at \href{https://github.com/FraGirla/MMSSL-for-CVD-Pred}{https://github.com/FraGirla/MMSSL-for-CVD-Pred}.

\section{Data}

Our analyses were performed on UK Biobank which is a prospective cohort study of over 500,000 individuals including their clinical data from across the United Kingdom \cite{sudlow2015uk,kappelhof2021evolutionary, velagapudi2021machine}. 


\textbf{Cardiac MR images:} The whole CMR image dataset is composed of 60,623 participants with dimensions of 208x208xZx50 with Z as a variable number of slices dependent on the size of the heart through 50 time points. To reduce data dimensionality, we selected the subset of CMR information that best characterizes a heartbeat. We used a pre-trained segmentation model from \cite{10.1007/978-3-030-00937-3_67,BAI201865} to label the left ventricle and calculate heart expansion based on the number of labeled pixels. This allowed us to determine the end-diastolic phase (maximum left ventricle expansion), the end-systolic phase (minimum expansion), and the mid-phase, defined as the midpoint between the two, as outlined in \cite{turgut2023unlocking}. Additionally, we reduced the 3D volume to 2D images by selecting the middle slice on the z-axis. This preprocessing approach extracts essential heartbeat information, reducing model complexity and improving speed.


\textbf{ECG:} We used 12-lead ECG data recorded for 10 seconds at rest with a sampling frequency of 500Hz and removed the measurement noise with 0.5 Hz high-pass bandwidth filter with order of 5, followed by powerline filtering \cite{Makowski2021neurokit}. The whole ECG dataset is composed of 50,531 participants each of them with 5000 time points and 12 leads. 


\textbf{Tabular data:} As the aim of this research is to take advantage of data that are easily available to make predictions on CVD, we integrate medical information as tabular features in the signal encoder that will be used to predict the downstream task. The list of 33 important clinical features included in this study are evaluated by medical experts and summarized a table in tabular data section in supplementary materials. Tabular dataset is composed of 45,257 participants with 33 features.
Categorical features were encoded using ordinal encoding and continuous features were normalized between 0 and 1.
 
We paired the ECG, CMRI, tabular features, and MI labels for multi-modal training. Some participants didn't have all three modalities available resulted in 45112 final dataset which we divided into 80\%, 10\%, and 10\% for training, validation, and test sets respectively. 

\section{Method}
We divided the training into four steps. First, we pre-trained the ECG encoder with an MAE that learns the latent space by reconstructing the ECG\cite{turgut2023unlocking}. For this part, we used all available ECGs. We employed the MAE pretraining for ECG as masked signal reconstruction is shown to be the most efficient self-supervised learning (SSL) method for ECG data \cite{yun2024automatic, sawano2024applying} and we think that the reconstruction task applied on all the leads is more effective in distinguishing between different ECG compared to the comparison between different augmentations made by SimCLR method. The image encoder with ResNet50 \cite{ResNet} backbone is separately pre-trained on CMR images using SimCLR \cite{SimCLR} loss to map two different views of the same image closer in the latent space. The selection of SimCLR is due to its demonstrated superior performance in pretraining on medical data compared to other self-supervised learning methods \cite{diong2024self,BestofBothWorlds}. Training the image encoder in advance helps transfer the knowledge from the image encoder to the signal encoder. In the next step, we use paired CMRI, ECG, and tabular data to improve the signal encoder of ECG. This is done using the CLIP loss \cite{CLIP} to transfer the learning from CMR images to ECG and tabular data. Eventually, we fine-tuned ECG and tabular encoder using a balanced dataset composed of all data from individuals having the disease and 
the same amount of randomly chosen healthy people.

\subsection{ECG masked autoencoder}

\begin{figure*}
  \centering
  \includegraphics[width=350pt]{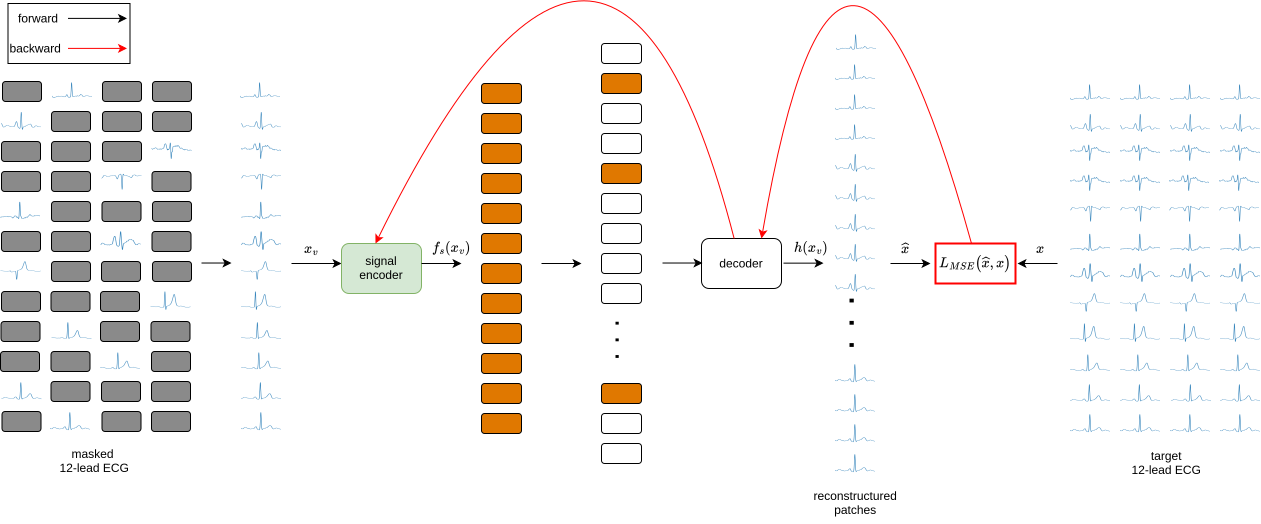}
  \captionsetup{skip=5pt}
  \caption{Visualization of the pre-training pipeline for ECG signals. The signal is divided into patches and a fraction of them is masked. The ECG encoder takes the visible patches 
  that will be reconstructed via decoder. MAE loss measures the distance between the reconstruction and the original signal. 
  A trained ECG encoder will be used in the next steps.}
  \label{fig:MAE}
\end{figure*}

Masked Autoencoder is a widely recognized self-supervised technique renowned for its ability to learn robust representations through its masking and reconstruction mechanism \cite{he2022masked}.
This approach aids in the acquisition of a robust ECG encoder capable of mitigating the redundancy inherent in ECG data, thereby enhancing sample discrimination. Additionally, this process facilitates the transfer of learning from CMRI to ECG and tabular encoders by leveraging an existing encoder. The pipeline for this step is summarized in Figure \ref{fig:MAE}.

\textbf{Masking:} The first step consists of dividing the whole 12-leads ECG signal into rectangular non-overlapping patches, masking a specified percentage of them using random sampling. While a mask ratio of 75\% is typically recommended \cite{he2022masked}, we have increased it to 80\% resulting in a larger effective receptive field that considers the ECG's redundancy such that the reconstruction task cannot be easily solved by interpolation.
 
We have as input a multi-channel time series $x \in \mathbb{R}^{C \times T}$, we divide the input in a flattened list of $N$ patches with size $D$ reshaping the original input to $x \in \mathbb{R}^{N \times D}$. Afterward, we randomly draw a binary mask on the patches for obtaining the visible patches $x_v = x\left[m\right] \in \mathbb{R}^{N_{v} \times D}$ and masked ones $x_m = x\left[1 - m\right] \in \mathbb{R}^{N_{m} \times D}$ such that $N = N_v + N_m$.

The backbone encoder is a ViT. We chose this model because we treat the ECG signal and its patches as an image that will be reconstructed thanks to its redundancy. Furthermore, it is a transformer-based technique that showed great reconstruction capabilities thanks to its ability to focus on relevant parts of the visible patches \cite{dosovitskiy2020image}. We haven't used a ResNet encoder as in the image encoding because we think that a reconstruction of the signal together with the attention mechanism are more beneficial for taking advantage of the periodicity and redundancy of the signal to learn quality ECGs embeddings. We apply this model to the visible patches. The encoder embeds patches by a linear projection with added positional embeddings and then processes the result via a series of Transformer blocks.


MAE decoder is given a set of encoded visible patches and mask tokens (introduced after the encoder) that represent a learned vector that indicates the presence of a missing patch to be predicted. Its task is to perform the prediction of pixels for each of the missing patches.
The decoder’s output $\hat{x} \in \mathbb{R}^{N \times D}$ is then reshaped to form a reconstructed image $\hat{x} \in \mathbb{R}^{C \times T}$.

MAE $h(.)$ consists of the composition of the signal encoder $f_s(.)$ and decoder $g(.)$ which, when provided with visible patches $x_s$, yielding the reconstructed signal $\hat{x}$.
\begin{equation}
    \hat{x} = h(x_v) = g(f_s(x_v))
\end{equation}


Further information about the augmentations and the training setting can be found in experimental setup section in supplementary materials.

\subsection{Image encoder pre-training}
\begin{figure*}
  \centering
  \includegraphics[width=350pt]{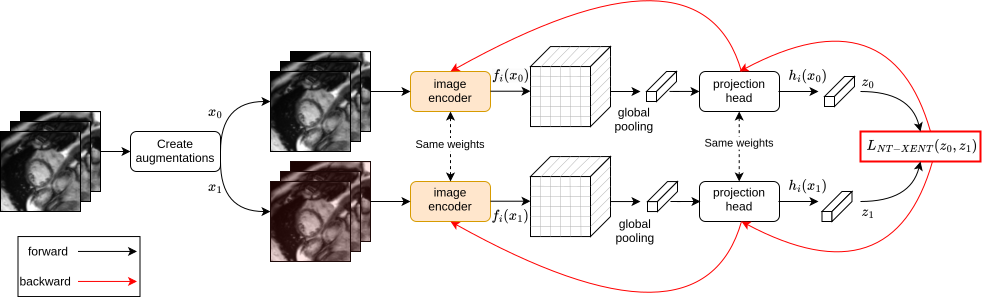}
  \captionsetup{skip=5pt}
  \caption{Visualization of the pre-training pipeline for CMRI. The image stack is augmented and given to image encoders. The resulting embeddings are then given to projection heads for optimizing the NT-Xent loss. The resulting image encoder will be later used for multi-modal self-supervised and uni-modal supervised training.}
  \label{fig:image}
\end{figure*}

We pre-trained the image encoder (Figure \ref{fig:image}) with SimCLR framework to learn representations by maximizing agreement between differently augmented views of the image via a contrastive loss in the latent space \cite{SimCLR}. The first step of this self-supervised technique consists of an input example $x$ and generating two image views $\tilde{x}_i$ and $\tilde{x}_j$ which we consider as positive pairs. 

After the augmentations, we use neural network $f(\cdot)$ that takes the two positive pairs and encodes them into representation embeddings. We use a ResNet50 as our backbone encoder since it has been proved that self-supervised techniques benefit more from bigger models compared to supervised techniques \cite{SimCLR}. We perform a forward pass for each of the two views and then make a global pooling. After the global pooling layer, we obtain $h_i = f(\tilde{x}_i) = \operatorname{ResNet}(\tilde{x}_i)$ where $h_i \in \mathbb{R}^d$ with $d=2048$.

We pass the resulting tensors to a projection head, denoted as $g(\cdot)$, responsible for mapping representations to the space where contrastive loss is applied. This projection head is removed during inference and replaced with a linear probing layer. Despite its seemingly counterintuitive nature, it has been theorized that this approach enhances the generalization of the learned embeddings, even in the absence of knowledge about the downstream task \cite{gupta2022understanding}. With the resulting projected embeddings, we calculate the NT-Xent loss \cite{SimCLR} \cite{MoCo} such that the image encoder maps two augmentations closer in the latent space. 

Our projection head consists of a multilayer perceptron (MLP) with a single hidden layer to obtain $z_i = g(h_i) = W^{(2)} \sigma(W^{(1)} h_i)$ where $\sigma$ is a ReLU non-linearity, $z_i$ represents the projected representation of the input, $W^{(1)} \in \mathbb{R}^{H \times I}$ and $W^{(2)} \in \mathbb{R}^{O \times H}$ are weight matrices associated with the hidden layer and projection layer respectively \cite{SimCLR}. In our setting the input of the projection head is an embedding of size $I = 2048$, we used a hidden layer size of $H = 2048$ and a projection size of $O = 2048$.


Given a set $\{\tilde{x}_k\}$ including a positive pair of examples $\tilde{x}_i$ and $\tilde{x}_j$, the contrastive prediction task aims to identify $\tilde{x}_j$ in $\left\{\tilde{\boldsymbol{x}}_k\right\}_{k \neq i}$ for a given $\tilde{x}_i$. 

We sample a mini-batch of $N$ samples, we augment them resulting in $2N$ data points. Then, given a positive pair we treat the other $2(N-1)$ within a mini-batch as negatives similar to Chen \etal \cite{chen2017sampling}.
The NT-Xent (the normalized temperature-scaled cross-entropy loss) loss is defined as:
\begin{equation}
    \ell_{i, j}=-\log \frac{\exp \left(\operatorname{sim}\left(\boldsymbol{z}_i, \boldsymbol{z}_j\right) / \tau\right)}{\sum_{k=1}^{2 N} \mathbbm{1}_{[k \neq i]} \exp \left(\operatorname{sim}\left(\boldsymbol{z}_i, \boldsymbol{z}_k\right) / \tau\right)}
\end{equation}
where $\mathbbm{1}_{[k \neq i]} \in\{0,1\}$ is an indicator function evaluating to $1$ if $k \neq i$, $\operatorname{sim}(\boldsymbol{u}, \boldsymbol{v})=\boldsymbol{u}^{\top} \boldsymbol{v} /\|\boldsymbol{u}\|\|\boldsymbol{v}\|$ denote the dot product between $\ell_2$ normalized $\boldsymbol{u}$ and $\boldsymbol{v}$ (i.e. cosine similarity) and $\tau$ denotes a temperature parameter \cite{SimCLR}.
Further information about the image augmentations and parameters used can be found in experimental setup section in supplementary materials.

\subsection{Multimodal Contrastive Learning}

\begin{figure*}
  \centering
  \includegraphics[width=350pt]{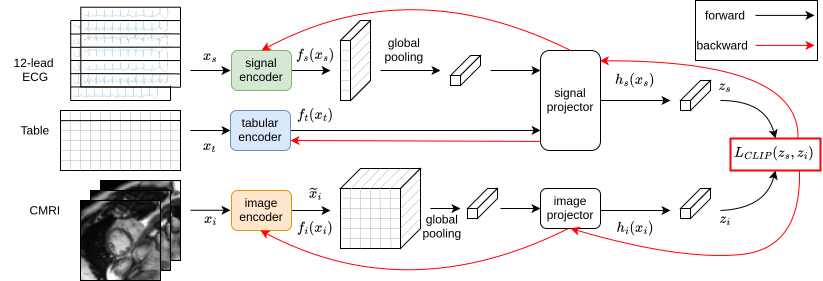}
  \captionsetup{skip=5pt}
  \caption{Visualization of the multi-modal pre-training pipeline. The three modalities are encoded taking advantage of the previous pretraining steps. The embeddings of ECG and tabular data are concatenated and optimized together. The projection head maps the two embeddings into the shared latent space where the CLIP loss is optimized.}
  \label{fig:multi}
\end{figure*}



We leverage the previously pre-trained encoders and introduce a third encoder specifically tailored to encode tabular features. By incorporating this additional encoder, we aim to capture the rich information embedded within the tabular data, complementing the representations learned from CMRI and ECG modalities as illustrated in Figure \ref{fig:multi}.


\textbf{Data augmentations:} We have the paired time series data $x_s$, tabular data $x_t$, and imaging data $x_i$ that are augmented and then processed by a signal encoder $f_s(.)$, a tabular encoder $f_t$ and an image encoder $f_i(.)$ to generate the embeddings for the downstream task.
As in the previous step, we take advantage of the same ECG and image augmentations. We augment the tabular features by corrupting a certain amount of randomly selected subjects' data following \cite{bahri2021scarf}. We used a corruption rate of 0.3. In the corruption process, the value of each feature is deliberately altered by sampling it with replacement from all observed values for that particular feature within the dataset. This technique is commonly referred to as sampling from the empirical marginal distribution. By introducing this corruption mechanism, the model is exposed to a range of possible variations in the data, enabling it to learn more robust and generalized representations. 


\textbf{Feature encoding:} Our tabular encoder is an MLP with one hidden layer of size 384 that generates embedding of size 384. The weights are initialized using Xavier uniform distribution \cite{glorot2010understanding}.
As both the image encoder and ECG encoder, we retained the architecture from the preceding step and utilized the pre-trained model weights. As the MAE encoder yielded embeddings for individual patches, we incorporated a global pooling layer to collapse the additional dimension leading to 1D embeddings.
As discussed in \cite{amal2022use}, we use a joint fusion strategy to merge ECG and tabular encodings. We will refer to the concatenated embeddings of the ECG encoder and tabular encoder as signal embedding. They represent the encoding of the modalities that are cheap and easily available and that will be contrasted to the CMRI modality which is expensive and more complex to acquire. We remark that only the signal embedding will be used in the prediction of the downstream task.


\textbf{Signal projection:} Both signal and CMRI embeddings are given to two SimCLR projection heads $g_s(.)$ and $g_i(.)$ to generate projections $z_s = g_s(f_s(x_s))$ and $z_i = g_i(f_i(x_i))$, respectively. The two projection heads have a hidden size of 256 and a final size of 128.
The resulting projections are $\ell_2$-normalized and mapped into a shared latent space.


\textbf{CLIP loss:} Furthermore, to facilitate the alignment of representations across all modalities and enhance the model's ability to learn informative features, we employ the CLIP loss function due to its effectiveness in encouraging the model to learn representations that can match input samples from different modalities \cite{radford2021learning}.

We obtain one loss for each modality:
\begin{equation}
    \mathcal{L}^{\operatorname{sig}}=\mathbb{E}_{p\left(x_s, x_i\right)} \mathbb{E}_{p\left(x_i^j\right)}\left[-\log \frac{\exp \left(z_s^{\top} z_i / \tau\right)}{\sum_{j \in B} \exp \left(z_s^{\top} z_i^j / \tau\right)}\right]
\end{equation}
\begin{equation}
    \mathcal{L}^{\text {img}}=\mathbb{E}_{p\left(x_i, x_s\right)} \mathbb{E}_{p\left(x_s^j\right)}\left[-\log \frac{\exp \left(z_i^{\top} z_s / \tau\right)}{\sum_{j \in B} \exp \left(z_i^{\top} z_s^j / \tau\right)}\right]
\end{equation}
then we combine it with a weighted average to obtain the total loss
\begin{equation}
    \mathcal{L}_{\text {CLIP }}=(1-\lambda) \mathcal{L}^{\text {sig}}+\lambda \mathcal{L}^{\text {img}}
\end{equation}
where $\tau$ is the temperature parameter and $\lambda$ is the parameter to balance the loss of the signal and imaging modalities.
Further information about the multimodal SSL step and parameters used can be found in experimental setup section in supplementary materials.


\subsection{Signal encoder finetuning}

\begin{figure*}
  \centering
  \includegraphics[width=350pt]{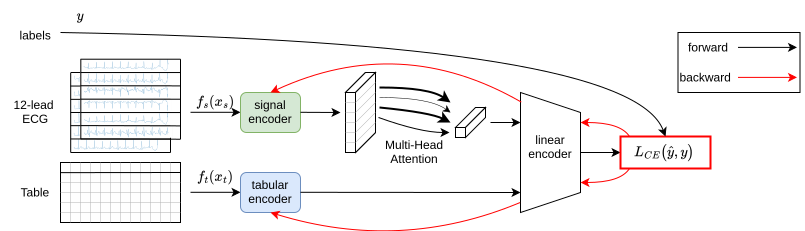}
  \captionsetup{skip=5pt}
  \caption{Pre-trained signal and tabular encoders are fine-tuned in a supervised manner including the labels for downstream task prediction.}
  \label{fig:fine-tune}
\end{figure*}

After enriching our embedding with multimodal contrastive learning, we fine-tune the signal encoder in a supervised fashion by training on a balanced dataset by optimizing the cross-entropy loss to predict better the downstream task as shown in Figure \ref{fig:fine-tune}. This balanced dataset comprises all patients with diagnosed disease and randomly chosen individuals from the healthy cohort. We evaluate the final model on the same validation and test set. As in \cite{turgut2023unlocking}, we replace the global pooling with a Multi-Head Attention \cite{vaswani2017attention} to map the encoder output to 1D embedding. 
Further information about the finetuning step and parameters used can be found in experimental setup section in supplementary materials.

To address the risk of overfitting in the fine-tuning step, our approach leverages a multimodal framework that has demonstrated strong performance even in low-data regimes, as evidenced by \cite{BestofBothWorlds} where they tested the performance of learned encoders in a low-data regime by sampling the fine-tuning dataset down to 10\% and 1\% of its original size. The results showed improved representation even with fewer samples compared to uni-modality methods which mitigates the concerns about overfitting with small fine-tuning dataset.

\section{Results}

In this section, we undertake a thorough comparison between different self-supervised techniques as well as a comparison between self-supervised and supervised learning methodologies for cardiovascular disease prediction. To address our imbalanced dataset issue, we employ evaluation metrics such as the receiver operating characteristic area under the curve (ROC AUC) and balanced accuracy which is a more reliable evaluation metric for imbalanced datasets.
Our prediction task specifically focuses on MI which is a critical medical condition that requires timely detection and intervention. Our task results in a binary classification for predicting the event based on the labels acquired from UK Biobank and shown as vascular disease in tabular features table in Supplementary materials.
Since annotated data is only used during the fine-tuning step, the encoder obtained at the end of multi-modal training remains flexible and can be repurposed for other cardiovascular disease predictions such as stroke. We have extended the method for predicting stroke and showed the results in supplementary materials.

\begin{table}[]
\centering
\captionsetup{skip=5pt}
\caption{Comparison of different diagnostic modalities and training strategy for CVD prediction. Columns indicate which pre-traing (pretr.)/training strategy is used. Best scores are in \textbf{BOLD} font. The second best is \underline{underlined}. Our approach outperforms all baseline models with regard to AUC and balanced accuracy (Bal. Acc) metrics.}
\label{tab:modalities}
\begin{tabular}{|c|c|c|c|c|c|c|c|}
\hline
 \textbf{SSL Modality} &
  \begin{tabular}[c]{@{}c@{}}MAE\\ pretr.\end{tabular} &
  \begin{tabular}[c]{@{}c@{}}Image \\ pretr.\end{tabular} &
  \begin{tabular}[c]{@{}c@{}}ECG\\ Train\end{tabular} &
  \begin{tabular}[c]{@{}c@{}}Tab.\\ Train\end{tabular} &
  \begin{tabular}[c]{@{}c@{}}CMRI\\ Train\end{tabular} &
  \textbf{\begin{tabular}[c]{@{}c@{}}AUC \\ {[}\%{]}\end{tabular}} &
  \textbf{\begin{tabular}[c]{@{}c@{}}Bal. \\ Acc \\ {[}\%{]}\end{tabular}} \\ \hline
Tabular                                                    & \xmark & \xmark  &  \xmark & \cmark  & \xmark & 0.62 & 0.5  \\ \hline
ECG                                                     & \xmark & \xmark  &  \cmark & \xmark & \xmark & 0.68 & 0.6  \\ \hline
ECG  & \cmark & \xmark  & \cmark  & \xmark & \xmark & 0.76  &  0.69  \\ \hline
 CMRI  & \xmark & \xmark  &  \xmark & \xmark & \cmark & 0.74  & 0.67   \\ \hline
\begin{tabular}[c]{@{}c@{}}ECG \\ (MMCL \cite{turgut2023unlocking})\end{tabular}   & \cmark & \cmark  &  \cmark & \xmark & \cmark & \underline{0.77} &  \underline{0.7}  \\ \hline
\begin{tabular}[c]{@{}c@{}}ECG + Tabular \\ (Ours)\end{tabular}    & \cmark & \cmark  &  \cmark & \cmark & \cmark & \textbf{0.80} & \textbf{0.71}  \\ \hline
\end{tabular}
\end{table}

To evaluate our model, we compare the performance across different modalities, namely CMR image, ECG signal, and tabular data from medical information. We assess our model's predictive accuracy when trained on individual modalities and combinations thereof. 
In examining the results of self-supervised training from table \ref{tab:modalities}, it becomes apparent that approaches solely utilizing tabular data or ECG signal lack the necessary granularity to effectively discriminate between the two classes, likely due to the inherent generality of the information provided by these modalities. However, when incorporating (MAE) pretraining, the performance of ECG-based models notably improves. This enhancement suggests that MAE pretraining contributes to bolstering the robustness and diversifying the training of the self-supervised method applied to ECG encoders, yielding more meaningful embeddings. Notably, utilizing only CMR images yields respectable performance, underscoring its status as the most informative modality for this predictive task. Moreover, the MMCL technique successfully transfers knowledge from CMR image to ECG, leading to a performance boost compared to uni-modal settings. In our method, we further extend this by adding the tabular modality, ultimately outperforming the previous settings. In summary, our approach leverages the strengths of all modalities, showcasing the effectiveness of multi-modal learning in enhancing predictive accuracy for CVD prediction.


We comprehensively compare our self-supervised method with supervised learning with Neural Network (NN) for cardiovascular disease prediction. We evaluate the performance of our model under both paradigms to assess the efficacy of self-supervised pre-training in enhancing predictive accuracy when the available annotated data is limited.
As a baseline, we train the NN \cite{erickson2020autogluon} in a supervised manner on tabular data, ECG signals, and a combination of them. We compared our method with simple NN supervised approaches due to the limited availability of annotated data, which restricted our ability to explore more advanced supervised methods. We extracted the features from ECG signal \cite{lubba2019catch22,ji2022time} for concatenating them with tabular features. As seen in table \ref{tab:self_supervised}, the performance of supervised learning on tabular data only and ECG signals only results in balanced accuracy of 54\% and 66\% respectively. Training the model on a combination of ECG signals and tabular information results in 65\% balanced accuracy which is likely due to the overfitting on the small task-specific data using only 1524 samples with balanced diseased and healthy individuals. However, our approach with pre-training the model on multiple modalities including images and fine-tuning on labeled data outperforms the supervised method by 7.6\% increase in balanced accuracy. 


\begin{table}[h]
\centering
\captionsetup{skip=5pt}
\caption{Comparison between self-supervised and supervised techniques}
\label{tab:self_supervised}
\begin{tabular}{|l|c|c|c|c|}
\hline
\textbf{Modality} & ECG & Tabular &\textbf{AUC [\%]} & \textbf{Balanced Acc [\%]} \\ \hline
Supervised NN    & \xmark & \cmark         & 0.55             & 0.54            \\ \hline
Supervised NN   & \cmark & \xmark          & 0.71             & 0.66            \\ \hline
Supervised NN    & \cmark & \cmark         & 0.70             & 0.65            \\ \hline
Ours   & \cmark & \cmark   & \textbf{0.80} & \textbf{0.71}             \\ \hline
\end{tabular}
\end{table}


\section{Conclusions}

This work presents a novel approach for enhancing the predictive performance for CVD diagnosis with limited data availability by leveraging transfer learning from CMR images to ECG signals and tabular clinical data. Our method, which incorporates self-supervised contrastive learning, offers a comprehensive framework for extracting meaningful embedding from diverse medical modalities and enhancing predictive capabilities by 7.6\% in balanced accuracy compared to supervised methods when using only ECG and tabular data for prediction.

Through extensive experimentation on UK Biobank dataset, we have demonstrated the effectiveness of our approach with significant improvement in predictive performance compared to supervised models. Moreover, we demonstrated that the steps in our pipeline progressively make the model learn richer embeddings and succeed in effectively capturing and aligning relevant features across modalities. This enables the model to learn informative representations that are robust and generalizable, ultimately leading to enhanced predictive accuracy for several downstream tasks.

Furthermore, the utilization of self-supervised learning techniques mitigates the need for large annotated datasets, making our approach scalable and applicable to real-world clinical settings where labeled data may be limited. We have demonstrated the potential for benefiting from other patient information such as text data with large language models.
In summary, our proposed method represents a promising avenue for advancing the field of personalized healthcare by learning holistic representations of the physiological state. 

Our work is limited by the reliance on the UK Biobank, which may not fully capture the diversity of broader populations due to its specific demographic characteristics. This could affect the model's generalizability and clinical effectiveness. Future studies should assess our approach using external data from more diverse populations to overcome these limitations and improve its global applicability. Additionally, exploring other SSL techniques might enhance patient discrimination and lead to more meaningful embeddings. Finally, assessing the limits of transfer learning from expensive to more affordable modalities could provide insights into which improvements could be made to improve these results.

\section*{Acknowledgement}
This research has been conducted using the UK Biobank resource under application number 81959. We also thank Prof. Samia Mora from Harvard Medical School and Brigham and Women’s Hospital for her invaluable medical insight and guidance, which greatly contributed to the success of our research. 
This project was supported by the grant \#2023-N-306 of the 1st Joint Call of the Swiss Data Science Center (SDSC) and the Strategic Focus Area “Personalized Health and Related Technologies (PHRT)” of the ETH Domain (Swiss Federal Institutes of Technology). N.D. is partially supported by the ETH AI Center postdoctoral fellowship. B.M. and N.D. acknowledge support of the Helmut-Horten-Foundation.

\bibliography{egbib}
\clearpage 
\section{Appendix}





\subsection{Experimental setup}

\subsubsection{MAE}

The specific ViT is equipped with 3 layers, and 6 projection heads that results in an embedding of size 384. We used a patch size of (1,100). Mean Squared Error (MSE) and the Normalized Correlation Coefficient (NCC) are used to evaluate the reconstruction performance where $\lambda_{MAE}$ is a parameter that weights the two components of the loss:
\begin{equation}
    \mathcal{L}_{MAE} = (1- \lambda_{MAE}) \mathcal{L}_{MSE} + \lambda_{MAE} \mathcal{L}_{NCC}
\end{equation}

Signal augmentations include random cropping at a ratio of 0.5, Fourier transform surrogate augmentation \cite{FTSurrogate} with a phase noise magnitude of 0.1, Gaussian noise with a sigma of 0.25, and a rescaling factor of 0.5.
We trained the masked autoencoder using AdamW optimizer \cite{loshchilov2019decoupled} with weight decay of 0.15, batch size of 128, base learning rate of 1e-5, and cosine annealing scheduler over 400 epochs with 10\% of warm-up epochs and a value of $\lambda_{MAE} = 0.1$

\subsubsection{Image Encoder}

We used different augmentation techniques such as random horizontal flips with a probability of 50\%, random rotations up to 45 degrees in the image, adding color noise to brightness, contrast, saturation, and a random resized crop of the image.

We trained the image encoder using the AdamW optimizer \cite{loshchilov2019decoupled} with weight decay of 1e-4, batch size of 512, base learning rate of 1e-4, and cosine annealing scheduler over 500 epochs with 10 warm-up epochs and the temperature parameter $\tau$ to 0.1.

\subsubsection{Multimodal SSL}

We trained the multimodal step using the AdamW optimizer \cite{loshchilov2019decoupled} with weight decay of 1e-4, batch size of 256, base learning rate of 1e-4, cosine annealing scheduler over 200 epochs with 10\% of warm-up epochs saving the checkpoint with the best loss. We set $\lambda$ to 0.5 to balance the components of the loss function and the temperature parameter $\tau$ to 0.1. We used both the global pooling layer and the attention pooling to help with the next step. We found similar results with both of them during the grid-search to find the right hyperparameters. 

\subsubsection{Finetune}
We trained the fine-tune step using the AdamW optimizer \cite{loshchilov2019decoupled} with weight decay of 1e-4, batch size of 64, base learning rate of 1e-5. We tried both cosine annealing scheduler over 400 epochs with 5\% of warm-up epochs, and a reduce LR on plateau scheduler, in the end we saw that the cosine annealing was the most consistent. We saved the best model for both schedulers according to the evaluation metric. 

\subsection{Extension to Stroke Disease}

We tried our method for predicting another cardiovascular disease (stroke). Our approach capitalizes on the available labeled stroke data to effectively leverage the learned representations' discriminative power. We find that the predictive performance, while promising, is not as robust as observed for myocardial infarction (MI). This discrepancy can be attributed to the limited size of the stroke dataset, which comprises only about half the number of instances compared to MI. By exploiting the informative labels associated with stroke instances, our method surpasses alternative approaches, and supervised methods achieving AUC improvement of 9.8\% and 21.8\% respectively (table \ref{tab:modalities_stroke} and \ref{tab:self_supervised_stroke}). This highlights the efficacy of our model in harnessing labeled data efficiently and underscores the quality of the trained embeddings, which proves useful even in scenarios with limited datasets, enhancing predictive performance for stroke disease prediction.


\begin{table}[]
\centering
\captionsetup{skip=5pt}
\caption{Comparison of different diagnostic modalities and training strategy for stroke prediction. Columns indicate which pre-train (pretr.)/training strategy is used. Best scores are in \textbf{BOLD} font. The second best is \underline{underlined}. Our approach outperforms all baseline models with regard to AUC and balanced accuracy (Bal. Acc) metrics.}
\label{tab:modalities_stroke}
\begin{tabular}{|c|c|c|c|c|c|c|c|}
\hline
 \textbf{SSL Modality} &
  \begin{tabular}[c]{@{}c@{}}MAE\\ pretr.\end{tabular} &
  \begin{tabular}[c]{@{}c@{}}Image \\ pretr.\end{tabular} &
  \begin{tabular}[c]{@{}c@{}}ECG\\ Train\end{tabular} &
  \begin{tabular}[c]{@{}c@{}}Tab.\\ Train\end{tabular} &
  \begin{tabular}[c]{@{}c@{}}CMRI\\ Train\end{tabular} &
  \textbf{\begin{tabular}[c]{@{}c@{}}AUC \\ {[}\%{]}\end{tabular}} &
  \textbf{\begin{tabular}[c]{@{}c@{}}Bal. \\ Acc \\ {[}\%{]}\end{tabular}} \\ \hline
Tabular                                                    & \xmark & \xmark  &  \xmark & \cmark  & \xmark & 0.59 & 0.5  \\ \hline
ECG                                                     & \xmark & \xmark  &  \cmark & \xmark & \xmark & 0.57 & 0.52  \\ \hline
ECG  & \cmark & \xmark  & \cmark  & \xmark & \xmark & \underline{0.63}  &  0.59  \\ \hline
 CMRI  & \xmark & \xmark  &  \xmark & \xmark & \cmark & 0.62  & \underline{0.60}   \\ \hline
\begin{tabular}[c]{@{}c@{}}ECG \\ (MMCL \cite{turgut2023unlocking})\end{tabular}   & \cmark & \cmark  &  \cmark & \xmark & \cmark & 0.61 &  0.57  \\ \hline
\begin{tabular}[c]{@{}c@{}}ECG + Tabular \\ (Ours)\end{tabular}    & \cmark & \cmark  &  \cmark & \cmark & \cmark & \textbf{0.67} & \textbf{0.62}  \\ \hline
\end{tabular}
\end{table}

\begin{table}[h]
\centering
\captionsetup{skip=5pt}
\caption{Comparison between self-supervised and supervised techniques. Best scores are in \textbf{BOLD} font. The second best is \underline{underlined}.}
\label{tab:self_supervised_stroke}
\begin{tabular}{|l|c|c|c|c|}
\hline
\textbf{Modality} & ECG & Tabular &\textbf{AUC [\%]} & \textbf{Balanced Acc [\%]} \\ \hline
Supervised NN    & \xmark & \cmark         & 0.55             & 0.52            \\ \hline
Supervised NN   & \cmark & \xmark          & 0.55             & 0.53            \\ \hline
Supervised NN    & \cmark & \cmark         & 0.55            & 0.53            \\ \hline
Ours   & \cmark & \cmark   & \textbf{0.67} & \textbf{0.62}             \\ \hline
\end{tabular}
\end{table}

\subsection{Tabular Data}

In table \ref{tab:tabular_data}, we present a comprehensive overview of the tabular data utilized in our study. This dataset encompasses a wide array of information, including demographics, comorbidities, and lifestyle factors. To handle missing tabular data, we imputed the categorical features with the most frequent ones and we used an iterative multivariate imputer for numerical features as a function of existing features over multiple imputation rounds. The last section also gives information about the cardiovascular diseases as labels that we want to predict. This tabular data will then be paired together with ECG and CMRI data so, as mentioned in Section 3, some tabular data could be missing.

\begin{table*}[ht]
\captionsetup{skip=5pt}
\caption{Tabular features for 45257 individuals}
\label{tab:tabular_data}
\centering
\begin{tabular}{|l|l|l|l|}
\hline
\textbf{Variable} & \textbf{Units} & \textbf{Descriptor} & \textbf{Missing} \\ \hline

\textbf{Demographics} & & & \\
\quad Age & Years (SD) & 64.6 (7.8) & 0\\  
\quad Waist circumference & cm (SD) & 88.7 (12.8) & 0\\
\quad Height & cm (SD) & 170.1 (9.4) & 0\\
\quad Weight & Kg (SD) & 76.1 (15.2) & 0\\
\quad BMI & Kg/$\text{m}^2$ M (SD) & 26.5 (4.4) & 0\\
\quad Sex & Female (\%) & 23375 (51.7) & 0\\
 
\textbf{Comorbidities} & & & \\
\quad Diabetes & Positive (\%) & 2542 (5.6) & 134\\
\quad Health rating & Good (\%) & 28532 (63.2) & 86\\ 
\quad Vascular heart problem & Positive (\%) & 2672 (5.9) & 14\\
\quad Stroke of father & Positive (\%) & 6345 (14) & 0\\
\quad Stroke of mother & Positive (\%) & 6389 (14.1) & 0\\
\quad Stroke of siblings & Positive (\%) & 1642 (3.6) & 0\\ 
\quad Breathe shortness & Yes (\%) & 3099 (6.9) & 645\\
\quad Anxiety visit & Yes (\%) & 13453 (29.9) & 256\\
\quad Chest pain & Yes (\%) & 4687 (10.4) & 365\\
\quad Stenosis & Positive (\%) & 131 (0.3) & 0\\
\quad Hypertension & Positive (\%) & 9798 (21.7) & 0\\
\quad Kidney disease & Positive (\%) & 1221 (2.7) & 0\\
\quad Dementia & Positive (\%) & 18 (0) & 0\\
\quad Thyrotoxicosis & Positive (\%) & 611 (1.4) & 0\\
\quad Migraine & Positive (\%) & 3344 (7.4) & 0\\
\quad Atrial fibrillation & Positive (\%) & 1427 (3.2) & 0\\
\quad Heart failure & Positive (\%) & 313 (0.7) & 0\\

\quad Embolism & Positive (\%) & 411 (0.9) & 0\\
\quad Deep-vein thrombosis & Years (SD) & 720 (1.6) & 35\\
 
\textbf{Lifestyle} & & & \\
\quad Smoke & Smoker (\%) & 923 (2.0) & 9\\
\quad Alcohol intake & Three or four times a week. (\%) & 12715 (28.1) & 19\\
\quad Diet salt & Never/rarely (\%) & 25820 (57.1) & 9\\
\quad TV Time & hour/day (SD) & 2.8 (1.6) & 125\\
\quad PC Time & hour/day (SD) & 1.5 (1.5) & 142\\
\quad Physical activity & number of days/week (SD) & 4.1 (2.2) & 1053\\ 
\quad Sleep duration & hours/day (SD) & 7.2 (1.1) & 128\\
\quad Coffee intake & cups/day (SD) & 2.0 (1.9) & 28\\

\textbf{Vascular Disease} & & & \\
\quad Stroke & Positive (\%) & 738 (1.6) & 0\\
\quad Myocardial infarction & Positive (\%) & 1399 (3.1) & 0\\

\hline

\end{tabular}
\end{table*}

\end{document}